# Deep Convolutions for In-Depth Automated Rock Typing

E.E. Baraboshkin[1], L.S. Ismailova[1], D.M. Orlov[1], E.A. Zhukovskaya[2], G.A. Kalmykov[3], O.V. Khotylev[4], E.Yu. Baraboshkin[5], D.A. Koroteev[1]

**Abstract**

The description of rocks is one of the most time-consuming tasks in the everyday work of a geologist, especially when very accurate description is required. We here present a method that reduces the time needed for accurate description of rocks, enabling the geologist to work more efficiently. We describe the application of methods based on color distribution analysis and feature extraction. Then we focus on a new approach, used by us, which is based on convolutional neural networks. We used several well-known neural network architectures (AlexNet, VGG, GoogLeNet, ResNet) and made a comparison of their performance. The precision of the algorithms is up to 95% on the validation set with GoogLeNet architecture. The best of the proposed algorithms can describe 50 m of full-size core in one minute.

## 1 Introduction

Mining and petroleum companies are keen to automate various processes in order to speed up exploration of mineral resources. This can be done by building large customized recommendation systems or expert systems, the use of mathematical methods, statistics. However, geological data are generally of uneven quality, due to varying behavior of well logging tools in different formations, differences between logging tools (sensitivity and precision), differences in geological descriptions and laboratory measurements, and it is difficult to build a sustainable mathematical model, which reacts appropriately to all of these data permutations.

In recent years, scientists in many fields have begun to use machine learning methods to analyze large volumes of data. Machine learning methods are based on linear and non-linear transformations of data. Machine learning can be used to design applications for self-driving cars, facial recognition, video classification. It is also applicable to geological problems. A key task when modeling depositional processes and sedimentary environments is facies analysis of a full-sized core. Creation of a good geological model depends on this task. The procedure has several stages, which makes it very time-consuming. First, the core must be described. The composition of the rock, its structure and texture and other characteristics must be identified. If the scale of description is large (1:10 cm and lower), it may take a geologist several weeks to complete this work. Then the description must be processed and analyzed along with the well logs, after which the facies can be tracked and the geological model can be built. The quality of a model depends on the accuracy of the description. If the model is built with low precision, then a significant part of the data may be misinterpreted (permeability and porosity distributions, estimates of petroleum reserve and much more). Various techniques are used to address these issues. Some of them are fairly obvious: to establish a compromise between precision of the description and quality of the geological model, or to automate the process of description. Automation usually depends on core image analyses. Machine learning can be applied to speed up routine image analyses. Several studies have been devoted to the automation of image descriptions.

[1] Skolkovo Institute of Science and Technology, Integrated Center for Hydrocarbon Recovery, Digital Petroleum, Nobel Street, Building 3, Moscow, 121205, Russian Federation, evgenii.baraboshkin@skoltech.ru
[2] Gazprom Neft, Science and Technology Center, 75-79 liter D, Moika River Embankment, St Petersburg, 190000, Russian Federation
[3] Lomonosov Moscow State University, Department of Geology and Geochemistry of Fossil Fuels, Leninskie Gory, 1, Moscow, 119991, Russian Federation
[4] Innopraktika, Lomonosovsky avenue, 27, Building 1.
[5] Lomonosov Moscow State University, Regional Geology and Earth History Department, Leninskie Gory, Building 1, Moscow, 119991, Russian Federation

The earliest known work on color distribution analysis is that of Prince and Chitale (2008). The authors used thresholding (separating by some value) of grayscale images in HSI (Hue, Saturation, Intensity) for pay estimation. They defined two classes – pay (sand) and non-pay (shale) – and tested the method on high-resolution images.

In 2011 Thomas et al. proposed a new method for automatic lithology classification. The method, called object-based image analysis, detects pixels which differ significantly from each other. Four lithotypes were used: sand, shale, carbonate and no core. First, different lithotypes are automatically separated from the core image by a threshold. After that, the nearest neighbor classifier (Arya et al., 1998) is applied to the small data (about 4.8 m of a full core with different lithology) to enable supervised learning. The nearest neighborhood classification is based on comparison of the whole dataset in n-dimensional space. Different classes were distinguished and placed at maximum distance from each other. The training dataset was created using samples of different lithotypes from the well. Subsequent classification was carried out on samples from the same well.

Thomas et al. (2011) applied the trained classifier to the whole dataset in order to classify lithotypes and the results were checked manually by an expert. Additional samples were added or removed from the nearest neighborhood model if the results were not reasonable. The method worked well when applied to one target, where all the data are from the same distribution (core image collection). In such cases accuracy was a high as 94.29%. Misclassification occurred in images with empty core or shaded areas.

An approach for improving color distribution analysis was offered in a series of works by Khasanov et al. (Khasanov, 2015, 2014, 2013; Khasanov et al., 2016; Postnikova et al., 2017). The key idea was to use the difference in lithology colors in HSI color space (limestones may be white or yellowish-white, sandstones yellowish white to green). Other petrophysical characteristics, such as oil saturation, porosity, permeability, cementation, can also be bounded through the color distribution. The approach requires choice of the right color distribution range for each type of characteristics (several hours of work). It is then necessary to determine which components are responsible for the color distribution (lithology, porosity/permeability). The authors proposed making a database of such color distributions and training an artificial neural network to use the color intensities to simplify the process.

Another method was based on the extraction of features using Principal Component Analysis (PCA) and color distribution analyses (Wieling, 2013). This research, which used various statistical tools, aimed to evaluate bedding direction, lithology, grain size, and permeability of a core that had been segmented to centimeter intervals. The detection was performed with an auto-covariance function. Information about each core image was represented in RGBD (Red, Green, Blue, Darkness) color space. The information was extracted by centered log ratio (CLR) transformation followed by PCA.

After PCA, information about bedding was extracted by correlation of the bedding image with two parallel lines imposed on the image, and then possible bedding directions were determined using k-means clustering. Lithology classification used a Multivariate Gaussian (MG) distribution model followed by Minimum Covariance Determinant and Quadratic Decision Boundary (QDB). The operations were carried out by linear regression of the Auto-Covariance properties for grain size and permeability estimation. Two lithology types were to be distinguished: sandstone and other rocks (coal, mudstone = shale (not to be confused with carbonates) and siltstone). Accuracy of the QDB classification model was 91% and 88% for the MG. So, the boundary between sandstone and other rocks was accurately defined. In order to distinguish the other rocks from each other, the contrast between siltstone and other coupled lithology classes was set by a linear decision boundary. But the results of this approach were disappointing, since siltstone, shale, and coal were all dark in color.

In works by Chatterjee et al. (Chatterjee, 2013; Chatterjee et al., 2010, 2008; Patel et al., 2017a, 2017b, 2016) crushed rocks from different mines were analyzed. Each rock was separately segmented from the image. Various features of texture and structure were extracted from the image by different methods. These features were reduced to a smaller dimensional space by principal component analyses (PCA) (until the 2010s) or by genetic algorithms (GA). In the most recent work (Patel et al., 2017a) there was no dimension reduction: 18 features of color distribution and intensity were used for the needs of online system development.

These features were placed into a multiclass support vector machine (SVM), which maps features that display non-linearity to linear space. After mapping, the non-linearity can be solved as a separate linear problem. Both approaches to separate the data were used: one-versus-all and one-versus-one. Precision of this method is between 90% and 99% (depending on the exact method used and the number of lithotypes).

Many other works devoted to ore classification problems have been thoroughly reviewed by Patel et al. (2017a). For example, Khorram et al. (2017) used SVM for classification of three different types of carbonates. The SVM method made the distinction with accuracy up to 89%.

Some other works describe the use of FMI (formation micro-imager) tools (Leal et al., 2018), OBMI (Oil-Base Micro-Imager) tools (Claverie et al., 2007; Knecht et al., 2004) and grain size analyses (Bukharev et al., 2018; Varfolomeev et al., 2016).

Recent works (Baraboshkin et al., 2018; Ivchenko et al., 2018) have shown that artificial neural networks can extract information from images and easily determine lithotypes. A development of this method is presented below. It can be used to describe a greater number of lithotypes and samples. Previous network architectures were unable to correctly classify new lithotypes, since they tended to underfit the data (were unable to generalize the data in order to make new predictions). We addressed this problem by investigating the abilities of various neural network architectures, constructing, training and testing the architectures. We gauged the performance of new lithotypes detection, and compared neural networks representational abilities.

## 2 Material and Methods

In previous work (Baraboshkin et al., 2018; Ivchenko et al., 2018) we showed how the identification of different rock properties can be automated using a convolutional neural network (CNN) (LeCun et al., 1989). We implemented new machine vision algorithms as the number of classes grew. The CNN can generalize information about images with many classes. The most common examples are ImageNet (Krizhevsky et al., 2012) (contains thousands of classes), Smiles (Hromada, 2010), flower classification (Nilsback and Zisserman, 2006) and Adience (gender and age recognition) (Eidinger et al., 2014).

To build the dataset, we collected 2000 core box images (approximately 2000 m of core). After preprocessing we had 20000 images with size 10x10cm (150 dpi). Each image was labeled with the corresponding lithotype. The lithotypes included: massive and laminated sandstone, limestone, shale (= argillite) and siltstone. The preprocessing algorithm was built to extract only core images, so there is no need to add a special class for it. Rocks were collected from different regions and formations in Russia, including Bazhenov, Abalak (Vasuganskaya and Georgievskaya), Vikulovskaya, Domanik and Achimov.

All computing was done on Python 3.6 (Van Rossum and Drake, 2011). For image operations, we used Python libraries: OpenCV (version 3.4.2) (Bradski, 2000) and NumPy (version 1.14.3) (Travis, 2006). To train and test the network we used backend Python libraries: Keras (version 2.2.4) (Chollet et al., 2015) with TensorFlow (version 1.10) (Martín Abadi, Ashish Agarwal, Paul Barham et al., 2015). For plotting the results, we used the Matplotlib library (version 2.2.2) (Hunter, 2007). For evaluation we used the Scikit-learn Python library (version 0.19.1) (Pedregosa et al., 2011). The Arkuda Skoltech cluster was used for training of the network.

### 2.1 Applied methods

#### 2.1.1 Convolution

Convolution is a process of matrix multiplication (element by element) followed by the sum of all elements of the matrix which fits into one number. The convolution in the artificial neural network (ANN) is equal to cross-correlation (Goodfellow et al., 2016):

$$S(i,j) = (K * I)(i,j) = \sum_m \sum_n I(i+m, j+n) K(m,n) \qquad (1),$$

where K stands for kernel, I for a two-dimensional image, and i, j, m, n are indexes of pixel positions in the kernel and image. The only difference is that the convolution is followed by transposition of the kernel, while the cross-correlation is not. This number goes into the new matrix as a pixel. Finally, some information about the image characteristics (Figure 1) is obtained, which may be information about shapes, lines, forms and other structural elements that exist in the image.

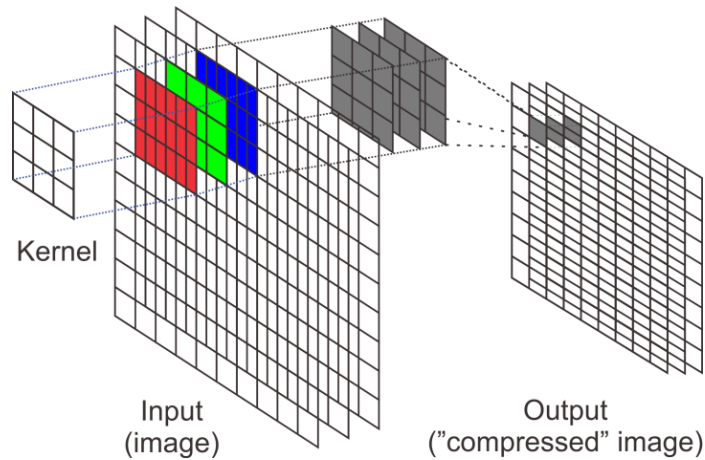

Figure 1. Schematic picture of a kernel of the convolutional neural network (CNN). The CNN usually contain hundreds of kernels (filters).

For example, we can apply a SobelX kernel (Freeman, 1990) for the image of laminated sandstone to extract information about lamination (figure 2).

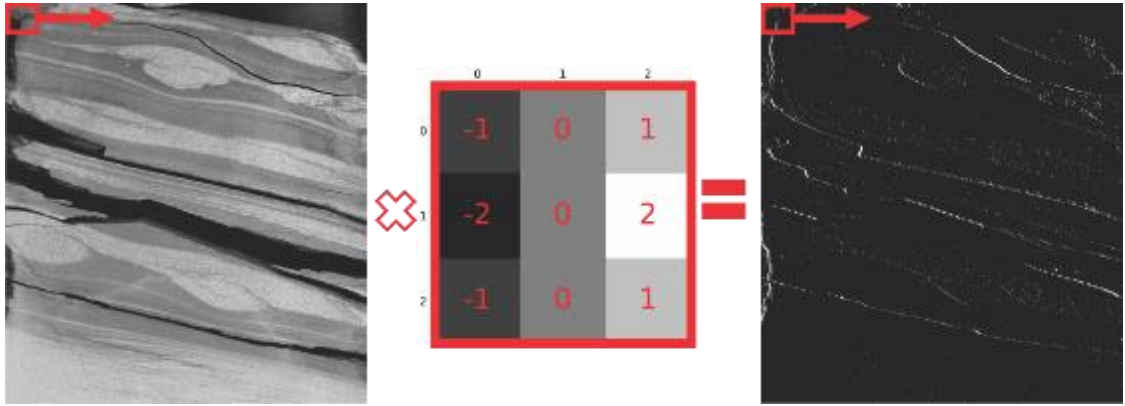

Figure 2. Example of the application of a SobelX kernel on a core image.

To prevent the reduction of spatial dimensions of the image (as the convolution transforms several pixels into one), additional pixels must be inserted (padded) into the image. The number of pixels to pad is calculated as follows:

$$P = \frac{F-1}{2} \qquad (2),$$

where P is padding size, F is filter size ("CS231n Convolutional Neural Networks for Visual Recognition Course," n.d.). Padding is usually carried out by zeros.

Such filters (kernels) can be trained by ANN (Goodfellow et al., 2016) to extract specific features from images for the best prediction. These filters are automatically trained to extract different features (e.g., grain size, lamination, color distribution, core condition) of the core images. Such ANN is called a convolutional neural network (CNN) (LeCun et al., 1990).

### 2.1.2 Convolutional neural networks

CNN is a specialized kind of neural network for processing data that has a known grid-like topology (Goodfellow et al., 2016). CNN consists of different layer types.

Layers in CNN have several parameters: depth, stride, and padding. Depth is the number of filters to train in the layer, stride is the number of pixels which the filter should process during convolution, and padding is a matter of whether or not padding should be carried out and what method should be used for it. Here a brief discussion of layer types can be seen. More concrete discussion of each layer type and its parameters can be found in Goodfellow et al. (2016).

The activation layer takes all the filters from the convolution layer and returns the same number of filters, which are binary thresholded by some value. The parameter of the activation layer is the activation function. There are several types of activation function. One of the most commonly used is the rectified linear unit (ReLu) (Nair and Hinton, 2010):

$$f(x) = max(0; x) \qquad (3).$$

Pooling layers take all the filters from the activation layer and return the same number of filters with reduced spatial dimensions (size and weight of filters). They are similar to convolutions. The difference between them is that the receptive field takes only one maximum value from the observed area.

A drop-out layer is required in order to regularize the network (not to produce large weights) and prevent overfitting. Overfitting is the state of the network when it can only predict labels from the training dataset with high accuracy. The drop-out layer randomly drops different connections in each filter. It gives the network much better generalization ability for predicting new data.

The fully-connected layer is connected to all neurons in the last convolution layer. It helps the network to make final decisions on how to label an image.

### 2.1.2.1 Artificial neural network optimization

Each machine learning algorithm should be optimized (i.e., the value of function *f(x)* should be maximized or minimized). When the function is minimized, it is called a loss function (Goodfellow et al., 2016). Gradient-based optimization is usually applied to minimize the loss function.

The major hyperparameter (i.e., parameter, which needs to be set by the researcher) for such methods is the learning rate (Goodfellow et al., 2016). The key point is that we never know whether the next step is good or bad for network performance. Learning rate optimization algorithms were invented to exclude human influence on the results of the training adaptive (Ruder, 2016). Such algorithms are commonly used in works appearing today.

We used the Adam (Adaptive Moment Estimation) optimization algorithm (Kingma and Ba, 2014) for the training of our network. It is based on the computing of adaptive learning rates for each parameter of the network. We also tried to apply other algorithms such as SGD (stochastic gradient descent) (Robbins and Monro, 1951), RMSprop (dividing the gradient by a running average of its recent magnitude) (Hinton et al., n.d.), but all of them gave less satisfactory network performance.

### 2.1.3 CNN architectures

Several components are needed to construct a simple CNN: a convolution layer (with weight, height, number of kernels (= depth) and other parameters), a fully connected layer (represents the number of classes) and, usually, a SoftMax function (S. Bridle, 1990), which is needed in order to calculate certainty in the prediction of the network.

Each of the pixels of this kernel (filter) is factually a node of a neural network. The number of nodes in one layer can be calculated as $N = h * w * k$ (4), where N is the number of nodes, h is the height of the layer output, w is weight of the output, and k is the number of kernels (depth) of the layer.

The architecture of the network is created from different layers, which were previously described. The architecture is the most valuable part of the CNN. Each architecture processes images in different ways. Each architecture works with different sizes and spatial dimensions of images due to different image transformations (Figure 3).

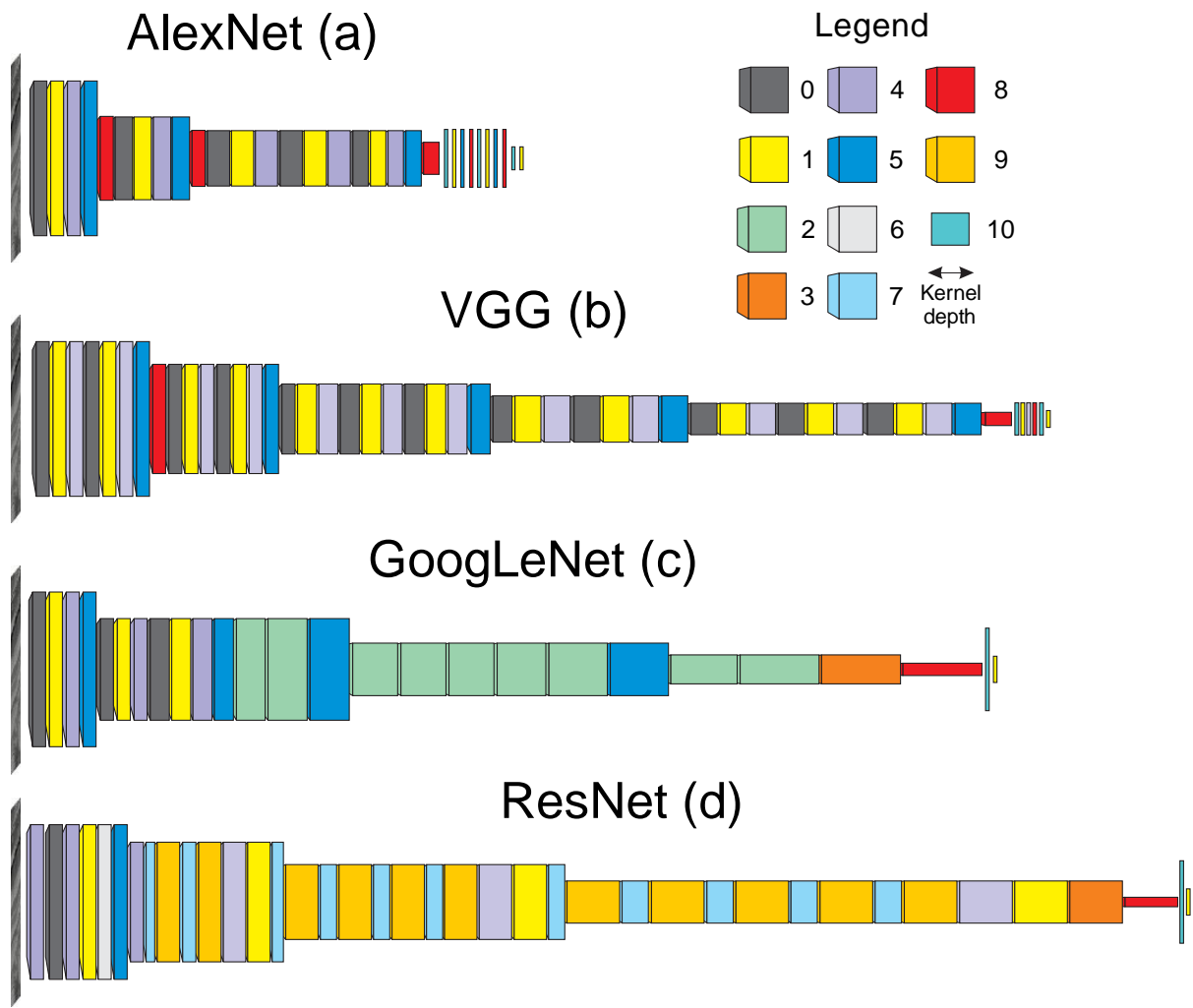

Figure 3. Simplified sketches of neural network architectures. Inception and residual module structures can be viewed in the detailed network maps in an online repository ("Supplementary materials to the article," n.d.). 0 – convolution layer, 1 – activation layer, 2 – Inception module, 3 – average pooling, 4 – batch normalization, 5 – max pooling, 6 – zero padding, 7 – Residual module, 8 – drop-out layer, 9 – layer composition, 10 – dense layer.

We implemented and trained several main architectures for image classification.

AlexNet architecture (Figure 3a) (Krizhevsky et al., 2012) consists of five convolutional layers and three fully-connected layers. AlexNet was the first architecture to be trained on several GPUs (previously networks were trained on CPUs or on one GPU). AlexNet was one of the first deep convolutional neural networks to achieve high accuracy on the ImageNet dataset (test error rate was 37.5%).

VGGNet (Figure 3b) (Simonyan and Zisserman, 2014) was proposed as the first very deep convolutional neural network. The aim was to evaluate how depth of the network influences its performance. The main conclusion was that the depth of the network does influence performance for the better. The maximum test error, which the authors obtained for ImageNet in this work, was 25.5%.

GoogLeNet (Figure 3c) (Szegedy et al., 2015; Zhang et al., 2017) was constructed and trained independently at the same time as VGGNet. Several new techniques were implemented to obtain high accuracy. The first technique, called "inception", concatenates different types of kernel outputs in one matrix. The second technique is to reduce the number of fully connected layers, enabling reduction of network storage space. The researchers also discovered that the width of the network significantly increased accuracy without overfitting. The work carried out in 2014 used 1x1 convolution layers to reduce the spatial dimension (the number of filters). The work from 2016 used the same approach to increase the width of the network. The test error was 6% for top-5 test error (top-1 accuracy error for inception-v3 in Zhang et al. (2016) was 37.5%).

The last architecture we used was ResNet (Figure 3d) (He et al., 2015). He et al. produced a new network training method which allowed better performance. The main idea was that the previous activation layers (a decision, which was made earlier) added to the new one (layer composition blocks, Figure 3). This allows better predictions as two decisions are taken into account for each layer. This method also easily overcomes various problems connected with deep network training, making it possible to easily train very deep networks (up to 1000 layers).

### 2.1.4 Image operation

Two color regimes were used: RGB and Grayscale. All images were normalized to a range of pixel intensities from 0 to 1 (the whole image was divided to 255). To train the classifier, all core images were cropped to 10x10 cm "samples" each 606x606 px in size (about 150 DPI). This size was chosen as the scale most frequently used by sedimentologists for core description (1:10 cm). All the images were resized to 227x227 px by the bilinear interpolation algorithm.

### 2.1.5 Image augmentation

It is a common case that datasets have different statistical distribution (Ostyakov and Nikolenko, 2019) which is called domain shift. That problem prevents the use of machine learning or statistical tools to wide range of data as due to domain shift the accuracy of models decrease. The augmentation is one of the ways to tackle this problem. Image augmentation is a useful tool which can produce many synthetic images from existing data. Such images will not be the same as the original dataset and will help the neural network to generalize information about the data. It can be done either one time as an oversampling of the dataset or each time during the training process. Effective data augmentation can improve the accuracy of the network (Bloice et al., 2017; DeVries and Taylor, 2017). We used different techniques for augmentation (Figure 4), which were available in OpenCV and NumPy: image rotation (orientations of core and lamination), brightness (as different photography settings) color manipulation (different composition of the rock) and random cropping (core plug extracted). The additional Albumentations library (Buslaev et al., 2018) was used for advanced augmentations, such as Gaussian Noise addition and blurring.

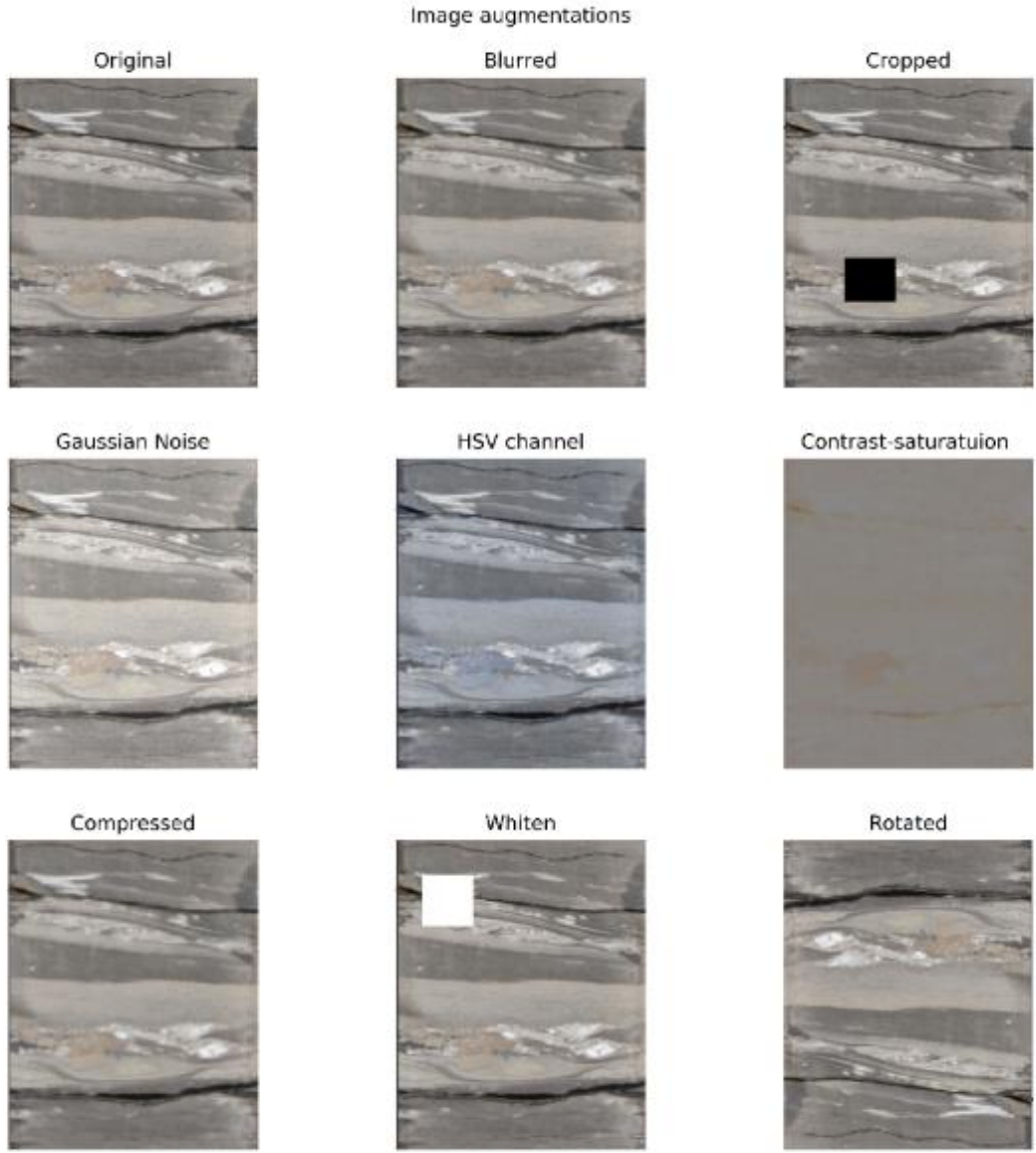

Figure 4. Different types of image augmentation.

### 2.1.6 Network performance evaluation

The data for training (85600 images, ~15000 per class after augmentation except for limestone, which had about 7000 images) were separated into training, validation and test sets. In order to evaluate the network performance around 110 images (50 for limestones) from each class were taken to validation and training. Eventually, both sets had 600 images. The network performance was evaluated after each training on the validation set. After training had been completed, the efficacy of the network was tested on new data (data not present in either the training or the test set). Finally, the network was tried on data from a new well, consisting of 440 samples.

We used several metrics to evaluate network performance (Fawcett, 2006; Metz, 1978): precision, recall, F$_\beta$-score, accuracy. As a good visualization tool of the network performance we've used confusion matrix (Fawcett, 2006; Metz, 1978). These techniques are standard for any machine learning tasks and are utilized for a better understanding of model performance. They provide valuable information where there are more than two classes with a different number of samples for each class. Accuracy is defined as $accuracy = \frac{TP+TN}{P+N}$ (Fawcett, 2006), where TP is true positive, TN is true negative predictions, P is true positive, N is true negative labels for prediction (the number of examples). Precision is a metric, which measures all TP against all true and false predictions of a class: $precision = \frac{TP}{TP+FP}$. Recall measures all TP predictions against all true labels: $recall = \frac{TP}{P}$. The F$_\beta$ measure shows the precision-recall relationship: $F_\beta = \frac{(1+\beta^2)*precision*recall}{(\beta^2*precision)+recall}$ with β=1, it serves for inclination toward precision (β=0.5) or recall (β=2). The confusion matrix unites all the information about classification in one table (Table 1).

Table 1. Example of a confusion matrix. TP – true positive classes, TN – true negative classes, 0 – shale, 1 – sandstone, 2 – limestone

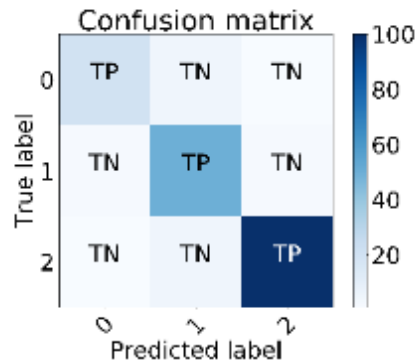

It can be easily seen in which classes predictions are correct and in which classes predictions are wrong. We used the Scikit-learn python library (Pedregosa et al., 2011) to generate the classification report.

## 2.2 Lithology classification

The main problem in lithology classification is that there are many different classifications of rocks (Wentworth, 1922). Practically every company creates its own classification to suit its needs. We used the standard initial origin-based classification (Bluman et al., 2015): sedimentary, igneous and metamorphic. We used simplified clastic sediment classification: sandstone (laminated and massive), siltstone, shale (=argillite). We distinguished carbonate rocks as a separate class. We did not take account of the origin of grain composition (a sandstone with a lahar is still clastic sediment and is still sandstone). This approach is quite similar that used by the British Geological Survey ("British Geological Survey (BGS) Rock Classification Scheme," n.d.).

# 3 Experiments

We trained our network using the Adam optimizer. The learning rate was set to 1e-3 (this rate worked well as the initial learning rate on all models). We used the polynomial learning rate decay

$$\alpha = \alpha_0 * \left(1 - ep/ep_{max}\right)^p \quad (5)$$

(Martín Abadi, Ashish Agarwal, Paul Barham et al., 2015; Rosebrock, 2017),

where α is a new learning rate, ep is the number of the current epoch, ep$_{max}$ is the maximum epoch number, and p is the power of the polynomial.

We also tried another method of learning rate manipulation: epoch-based learning rate decay. When the epoch reaches a certain number, the learning rate changes by one decimal point. This method worked better than polynomial learning rate decay, but can sometimes lead to spikes, as seen on the plots (Figure 5, Smith, 2017).

We also tried to apply fine-tuning (Chollet, 2017), a technique that uses models which have been trained to solve some other task. We used networks trained on an ImageNet dataset. The networks were unable to generalize the data, even after the activation of all layers for training. The randomly initialized networks converged much faster.

Our experiments with all of the above-mentioned architectures enabled us to generate a technique that gave excellent results on new data without any overfitting. Each network was trained for one day on 1 Nvidia k80 GPU.

## 3.1 Results
### 3.1.1 Training

The experiments using previously described architectures for a new set of lithotypes gave high levels of accuracy. In most cases only 20 epochs were needed to obtain low classification error (5-10%) on the training dataset (Figure 5). All networks showed good results during training.

AlexNet                                   ResNet

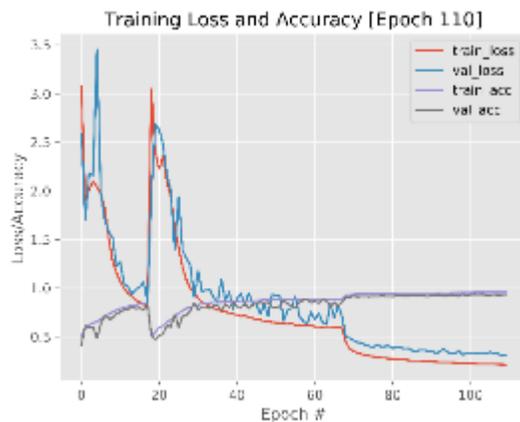 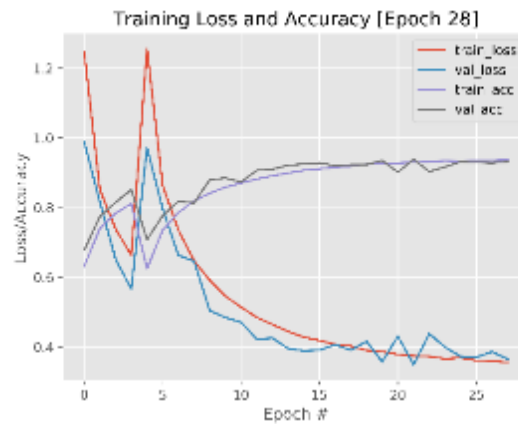

Figure 5. Evaluation of network performance. Training loss and accuracy plots for the best experiment.

Confusion matrix analysis (Table 2) shows that the laminated sandstone class is sometimes confused with massive sandstone and siltstone. Some of the laminated sandstones may indeed be referred to each of these classes by a geologist. Some of the laminated sandstones have very fine-grained lamination which can be easily referred to siltstones. Probably for the same reasons, some of the fine-grained massive sandstones were misinterpreted as siltstones.

Table 2. Confusion matrix analysis and performance statistics for the trained networks. Annotation for the confusion matrix labels: 0 – argillite, 1 – granite, 2 – limestone, 3 – laminated sandstone, 4 – massive sandstone, 5 – siltstone. The color bar indicates the number of samples.

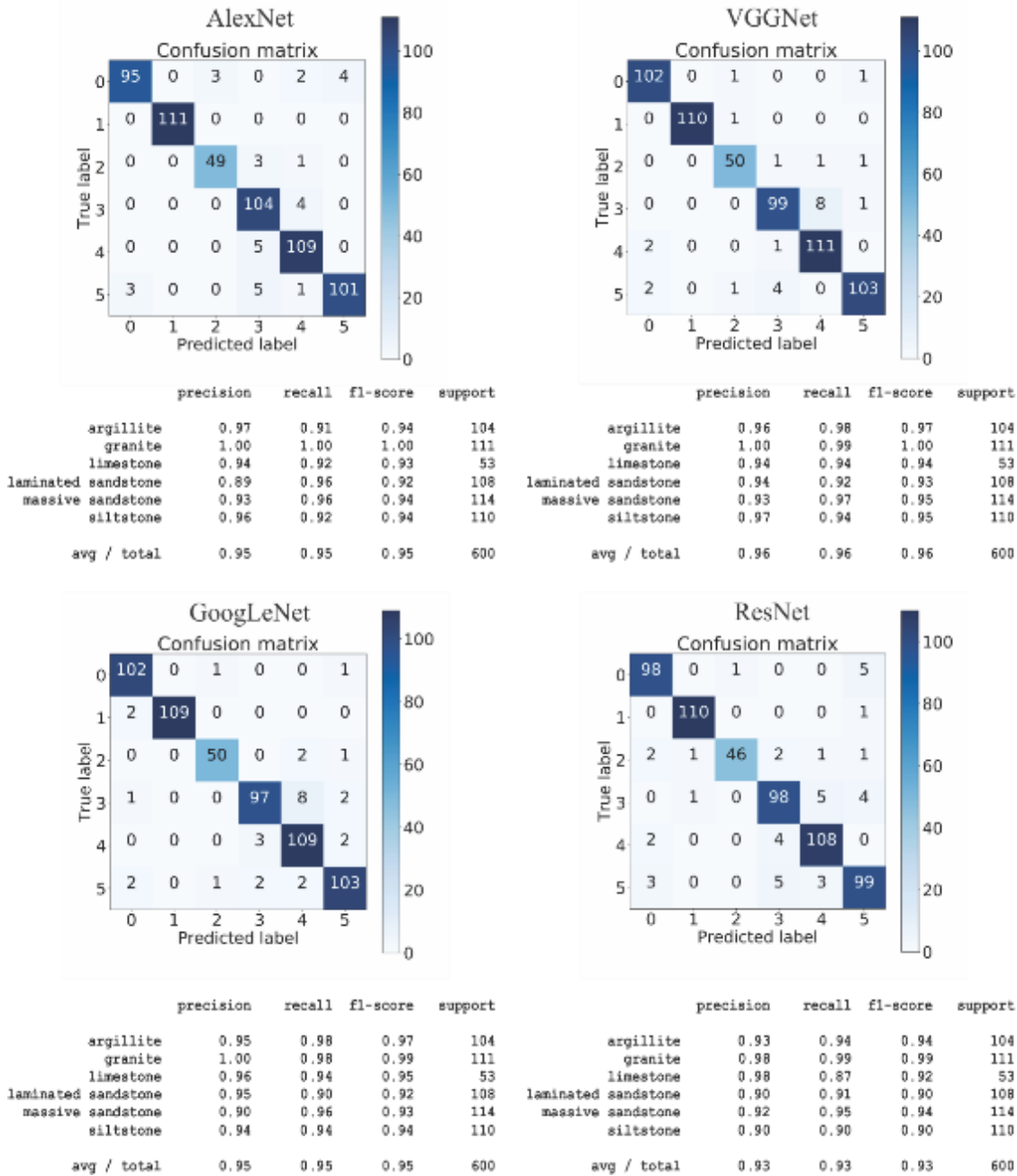

### 3.1.2 Neural network representation

When artificial neural networks first appeared, they were generally perceived as "black boxes". ANN representation techniques have been created since 2012 (Chollet, 2017) in order to "unbox" them. The image is passed through the network and information about it is extracted as a feature map from certain layers. This lets us understand how the system interprets the images and why it makes mistakes. The first network that was tested (Baraboshkin et al., 2018; Ivchenko et al., 2018) generated different structures (lamination or grains) on each layer. Different layers of the new architectures interpret the images in different ways.

Part of the feature map is represented in Figure 6; each layer name can be found in the large schemes of ANNs in the appendix ("Supplementary materials to the article," n.d.). A set of four trained filters (out of at least 32) from each layer are presented for demonstration. Each lithotype raises different activations for the set of filters, as can be seen in Figure 6. In each feature map, different areas are highlighted with different weight for different lithotypes. So the filters capture different characteristics of each lithotype.

Local interpretable model-agnostic explanations (LIME), proposed by Ribeiro et al. (2016), offer another way of understanding which areas of the image are used to make a prediction. This method lets us see the areas of an image, which have greatest influence on the prediction results (Figure 7).

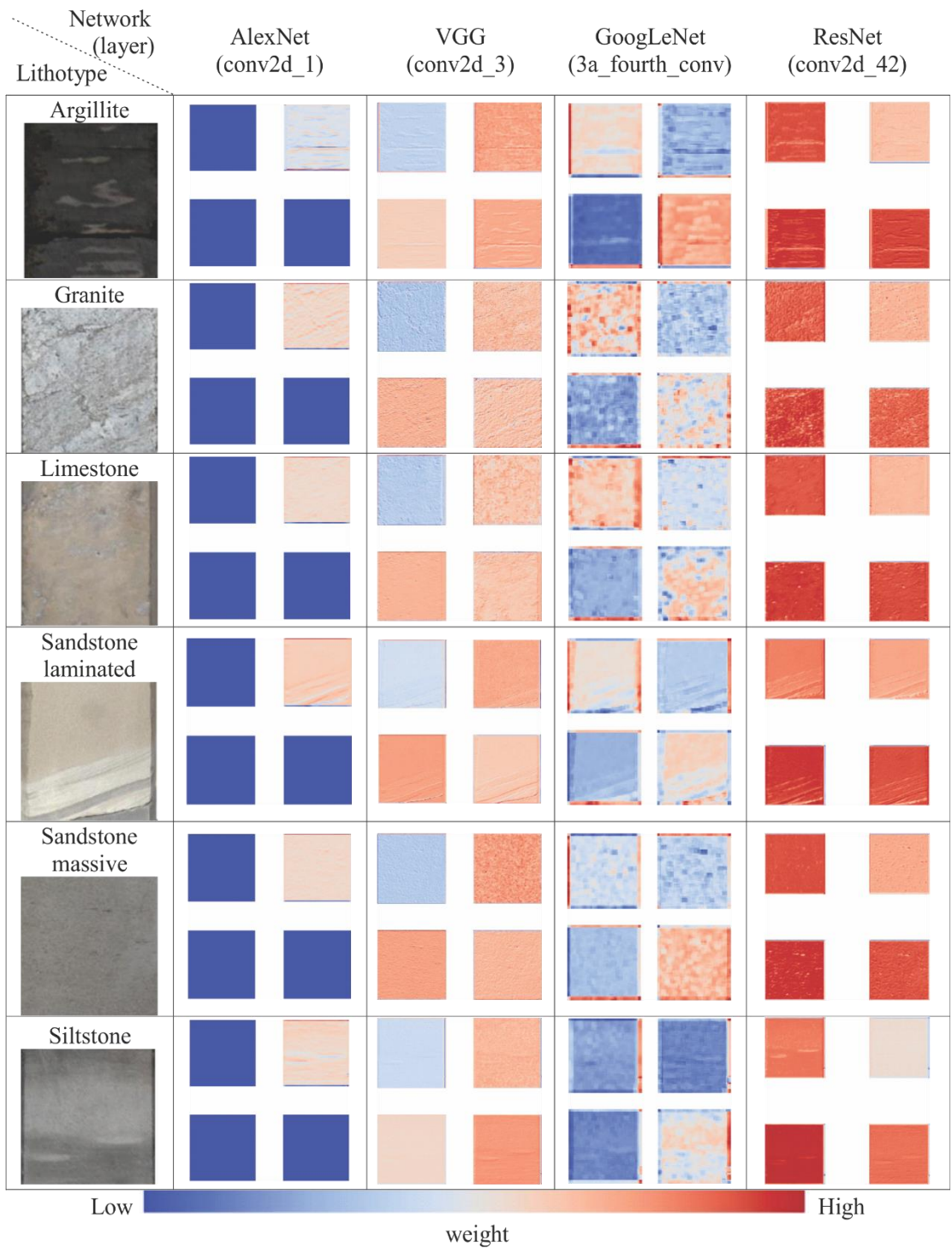

Figure 6. Maps of layer features extracted by the neural network, applied to our data after completion of training. Difference in the sizes of the maps is due to different sizes of filters

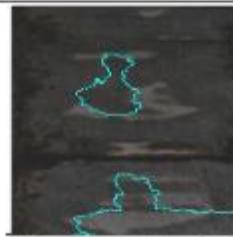

Figure 7. Areas of the image (outlined in color), which enable the network to predict the lithotype.

A set of feature maps from each neural network was applied to each lithotype. Only AlexNet activated a limited number of filters from a large number in all lithotypes (only one filter out of four was activated in Figure 6).

As can be seen, in the other architectures the filters have the same weight sign for most of the lithotypes, but each filter selects particular features for each lithotype. ResNet and VGGNet concentrate on the grain size of the lithotype, as each filter emphasizes the granularity of an image. ResNet does not extract information from images when the size of grains is small (the siltstone and argillite images do not have large weight fluctuation). However, VGGNet attempts to extract information in all cases.

GoogLeNet activations concentrate more on texture classification. The granularity of intermediate layers is small; it can be seen only in the massive sandstone and granite.

In all cases the networks detect the lithotype based on several image regions of different sizes (Figure 7), which may sometimes lead to misclassification. In presented images most of them were predicted correctly except the siltstone lithotype. AlexNet and GoogLeNet classified the image as massive sandstone. That is happened due to most of examined patches of image are similar to massive sandstones, which can be found in the same figure. ResNet classified image as laminated sandstone as some patches indeed had lamina and others were similar to sandstone. VGGNet classified image correctly as siltstone probably because it found same patches previously in training set. The VGGNet made a misclassification error while predicting the granite image as laminated sandstone. The marked matches contain some lamina and texture similar to sandstone. Same features can be found while comparing extracted features in Figure 6. Extracted features by VGGNet from granite and laminated sandstone are similar (the same sizes of granularity and lamination).

### 3.2 Discussion

In order to test the networks, we built a new data set and compared the algorithm performance with the opinion of an expert. Lithology of 44 m of new core was described. Each network predicted the core with high accuracy (Figure 9). No special equipment is needed to run the trained network, which operates on a CPU. A GPU speeds up action of the algorithm. The information processing speed is around 50 m per minute on a GPU and around 25 m per minute on a CPU, depending to a large extent on characteristics of the workstation.

The results of the comparison are shown in Figure 9 and Table 3. The conclusions concerning network performance are similar to what was stated in Section 3.1.1 above. For the final test 440 samples were provided (44 m of new core). As the new dataset may differ from the initial statistical image distribution (the same as lithotype descriptions which may be made by other geologist) due to domain shift problem the performance of the network may drop in different cases. Most of the samples were recognized correctly by all neural networks. AlexNet sometimes mixes shales and siltstones. There were some serious problems when oil-saturated sandstone images were used in processing: 21 samples were wrongly predicted as limestone instead of sandstone. The same problem occurred for laminated sandstones (15 of them were wrongly referred to limestones). As can be seen (Figure 8) the limestones from the training dataset are like the oil-saturated sandstones from the new data. So, the problem with confusion of these two lithotypes can be solved by the provision of more images of oil-saturated sandstones. Also, the sandstones can be additionally augmented by the application of yellowish-white filters to simulate the saturation Ten bioturbated samples, which should be referred to siltstones were detected as granites as they were similar to granites in the training datasets (Figure 8). VGGNet is more successful in identifying sediments with bioturbation structures. Only one sample and 1 argillite with 2 siltstones were predicted as granite. VGGNet, like AlexNet, fails to predict oil-saturated sandstone and laminated sandstones and shales (55 oil-saturated samples and 1 shale sample were predicted as limestones).

GoogleNet failed to detect 48 samples as sandstone and 2 samples as shales (both were detected as limestones). The network understood 6 bioturbated samples as granites.

ResNet made the least number of errors (argillite (1), laminated sandstones (6), massive sandstones (5), and siltstones (7) were detected as limestones, and three agillites and one siltstone were detected as granites). Other mistakes are connected to the near-boundary lithotype samples which can be detected as sandstones or siltstones. (the problem was already discussed in Section 3.1.1, above)

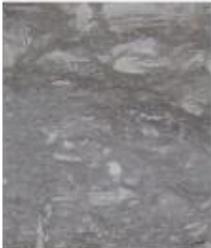

Figure 8. Misinterpreted lithotypes. The "New data" and "True labels" columns represent the images and their labels from the new well. "Predicted label" is the prediction produced by the CNN. The "Similar examples..." column contains images from the predicted classes of a training dataset.

After analysis of the results, we concluded that most of the rocks that were wrongly identified were, in fact, highly complex (as in Figure 9, Figure 8) and the mistakes committed by the networks are understandable. One of the most problematic classes is limestone. This is probably because of the small number of images in this class (7000 examples instead of 15000 for other classes). Another problem is that, in many instances, the image contains near-classification border grains and can be described by both lithotypes: sandstone and siltstone. The maximum precision in detection of multiple classes was 72% (GoogLeNet) and best performance for recall was 60% (ResNet). As regards speed of work, 70% of the core (about 30 m) was correctly identified in one minute. We thus have a very useful tool, which can help geologists to describe core lithotypes much faster than before.

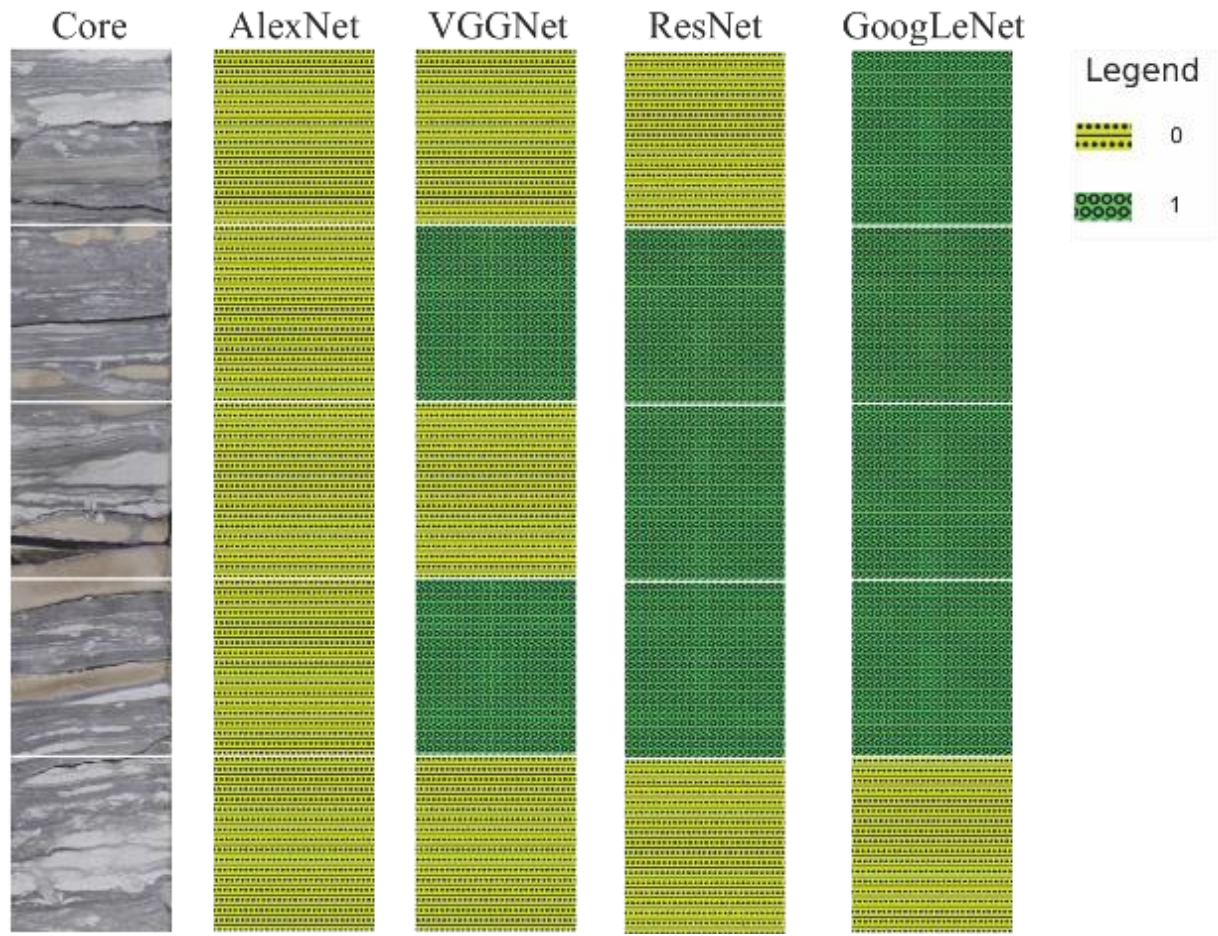

Figure 9. Comparison of performance by the networks on new data. The size of the sample is indicated by white spaces. Core width is 10 cm. 0 – laminated sandstone, 1 - siltstone

Table 3. Confusion matrix analysis and performance statistics for an expert description as True labels and results of CNN predictions as Predicted labels. The data used to evaluate CNN is from a new core (440 samples). Annotation for the confusion matrix labels: 0 – shale, 1 – granite, 2 – limestone, 3 – laminated sandstone, 4 – massive sandstone, 5 – siltstone. The color bar indicates the number of samples.

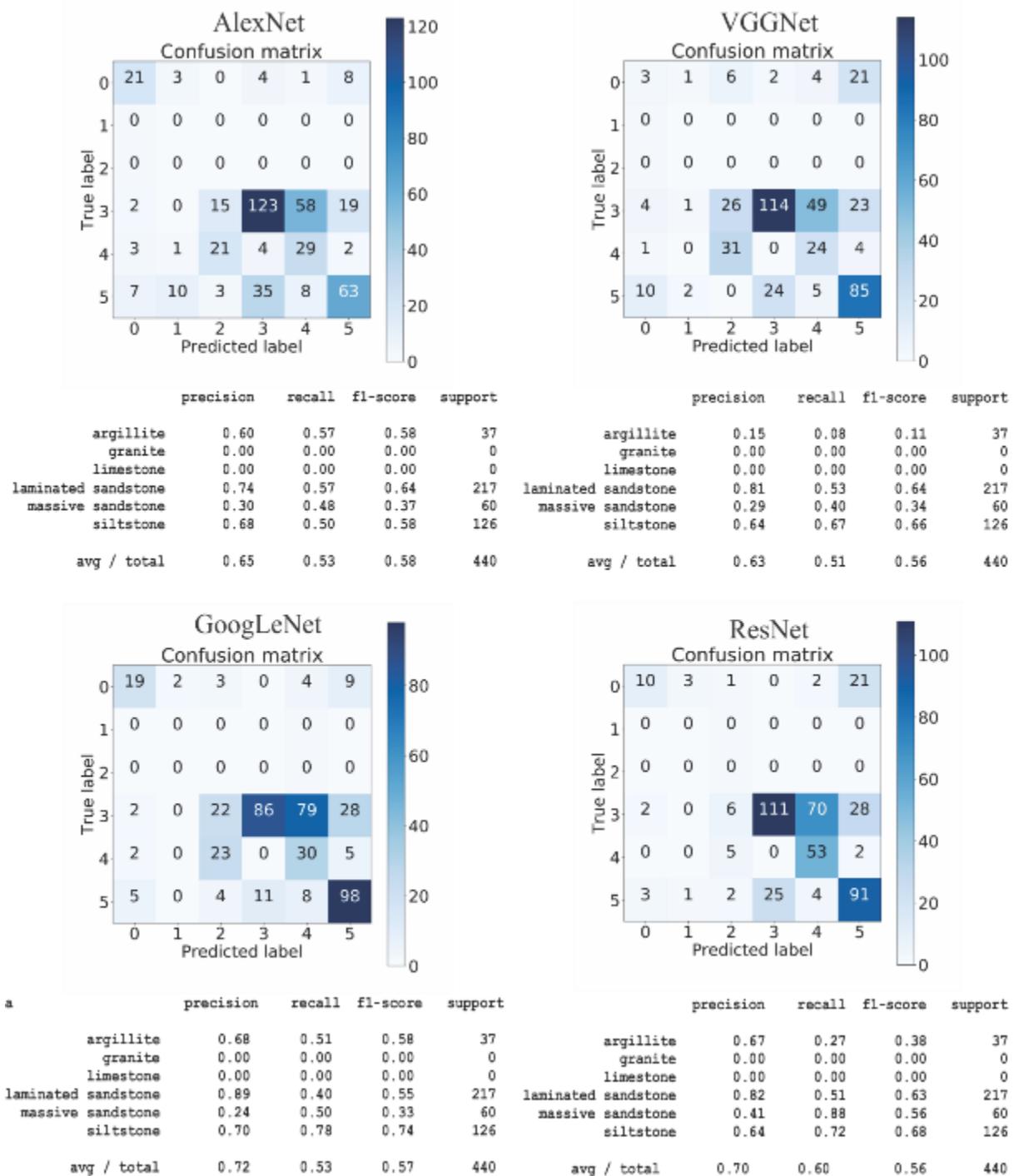

The work that has been done offers a means of optimizing and considerably reducing the time required for core lithotype description. Further research will aim to increase precision and recall. One way of improving the classification results would be to train a model simultaneously with regular images and images taken in ultraviolet light, and using actual well-log, core samples and core-log information. There are several problems, mainly connected with the depth correlation of each database, which need to be solved. We will further expand the lithotype database to understand the limits of the method. We also intend to achieve full description of the core, including structure and texture classification and other characteristics. The problem of persistence of several classes in one image could be solved using segmentation algorithms. This approach needs to be investigated.

### 3.3 Conclusion

We have presented a new approach to the automated classification of rocks, which can help geologists to reduce the time spent on core description. The available methods for automatic classification were described and a new application of the known convolutional neural network method was proposed. Different types of convolutional neural network architectures were studied and compared.

The best ANN architectures are GoogLeNet and ResNet. They demonstrate high levels of precision (72% and 70%) and recall (53% and 60%) with average f1 scores of 0.57 and 0.56, respectively.

We trained the automated classification system on 20000 image samples from over 10 oil and gas fields with different stratigraphy, transport and storage properties. To overcome the problem of domain shift we've made augmentation of images. We carried out additional verification of the model on a new dataset.

We also discovered several limitations to CNN classification. Firstly, some lithotypes can produce comparable structures, such as granites with fluctuation structure and bioturbated sediment. Secondly, images may contain two or three lithotypes, which leads to misinterpretation when they are compared with the expert description. Finally, some lithotypes may be referred to several types due to fine grain size or similarity of structures.

In future work, we plan to go further in the description of rocks and to attempt automation of texture-structure classification, enabling higher resolution of the rock typing and pixel-to-pixel lithotype classification. Measurements of physical properties along the core and simultaneous analysis will be conducted during prediction in order to improve network performance.

## 4 Acknowledgments

The authors thank Alexander Vladimirovich Ivchenko (MIPT) for his advice during preparation of the paper. The authors are also grateful for the use of the Skoltech CDISE HPC Arkuda and Pardus cluster, which made it possible to obtain the results presented in this paper.

## 5 Funding

This research did not receive any specific grant from funding agencies in the public, commercial, or not-for-profit sectors.

## 6 Data Availability

Datasets related to this article can be found at open-source online data repositories: http://www.bgs.ac.uk/opengeoscience/photos.html (hosted by the British Geological Survey), http://geocollections.info/ (National Geological Collection of Estonia) and https://nopims.dmp.wa.gov.au/Nopims/ (Geoscience Australia).

# 7 Computer Code Availability

No new software/script was developed during the research work.